\documentclass[aps,pre,twocolumn]{revtex4-1}
\usepackage{graphicx}
\usepackage{amsmath}
\usepackage{amssymb}
\usepackage{psfrag}
\usepackage{tabularx}
\usepackage{setspace}
\usepackage{url}

\usepackage{xcolor}
\newcommand{\be}{\begin{equation}}
\newcommand{\ee}{\end{equation}}

\newcommand{\bea}{\begin{eqnarray}}
\newcommand{\eea}{\end{eqnarray}}

\begin{document}

\title{Dialogue Transformers}

	\author{Vladimir Vlasov}
	\email{vladimir@rasa.com}
	\affiliation{Rasa}
	\author{Johannes E.\ M.\ Mosig}
	\email{j.mosig@rasa.com}
	\affiliation{Rasa}
	\author{Alan Nichol}
	\email{alan@rasa.com}
	\affiliation{Rasa}
	%\date{}
% \author{%
%   Vladimir Vlasov \\
%   Rasa \\
%   \texttt{vladimir@rasa.com} \\
%   % examples of more authors
%   \And
%   Johannes E.\ M.\ Mosig \\
%   Rasa \\
%   \texttt{j.mosig@rasa.com} \\
%   \AND
%   Alan Nichol \\
%   Rasa \\
%   \texttt{alan@rasa.com} \\
% }

\begin{abstract}
We introduce a dialogue policy based on a transformer architecture~\citep{vaswani2017attention}, where the self-attention mechanism operates over the sequence of dialogue turns.
Recent work has used hierarchical recurrent neural networks to encode multiple utterances in a dialogue context, but we argue that a pure self-attention mechanism is more suitable.
By default, an RNN assumes that every item in a sequence is relevant for producing an encoding of the full sequence, but a single conversation can consist of multiple overlapping discourse segments as speakers interleave multiple topics.
A transformer picks which turns to include in its encoding of the current dialogue state, and is naturally suited to selectively ignoring or attending to dialogue history. 
We compare the performance of the Transformer Embedding Dialogue (TED) policy to an LSTM and to the REDP~\citep{vlasov2018few}, which was specifically designed to overcome this limitation of RNNs.
\end{abstract}

\maketitle

\section{Introduction}

Conversational AI assistants promise to help users achieve a task through natural language. 
Interpreting simple instructions like \emph{please turn on the lights} is relatively straightforward, but to handle more complex tasks, these systems must be able to engage in multi-turn conversations.

The goal of this paper is to show that the transformer architecture~\citep{vaswani2017attention} is more suitable for modeling multi-turn conversations than the commonly used recurrent models.
To compare the basic mechanisms that are at the heart of the sequence encoding we intentionally choose simple architectures. 
The proposed TED architecture should be thought of as a candidate building block for use in developing state-of-the-art architectures in various dialogue tasks.

Not every utterance in a conversation has to be a response to the most recent utterance by the other party. 
Groz and Sidner~\citep{grosz1986attention} consider conversations as an interleaved set of \emph{discourse segments}, where a discourse segment (or topic) is a set of utterances that directly respond to each other. 
These sequences of turns may not directly follow one another in the conversation.
An intuitive example of this is the need for sub-dialogues in task-oriented dialogue systems. Consider this conversation:

\small
\begin{verbatim}
 BOT: Your total is $15.50 - shall I
      charge the card you used last time?
 USER: Do I still have credit on my
       account from that refund I got?
 BOT: Yes, your account is $10 in credit.
 USER: Ok, great.
 BOT: Shall I place the order?
 USER: Yes.
 BOT: Done. You should have your items tomorrow.
\end{verbatim}
\normalsize

\paragraph{Dialogue Stacks}    
The assistant's question \emph{Shall I place the order?} prompts the return to the task at hand: completing a purchase.
One model is to think of these sub-dialogues as existing on a stack, where new topics are pushed on to the stack when they are introduced and popped off the stack once concluded.

In the 1980s, Groz and Sidner~\citep{grosz1986attention} argued for representing dialogue history as a stack of topics, and later the RavenClaw~\citep{bohus2009ravenclaw} dialogue system implemented a dialogue stack for the specific purpose of handling sub-dialogues.
While a stack naturally allows for sub-dialogues to be handled and concluded, the strict structure of a stack is also limiting. 
The authors of RavenClaw argue for explicitly tracking topics to enable the contextual interpretation of the user intents. 
However, once a topic has been popped from the dialogue stack, it is no longer available to provide this context.
In the example above, the user might follow up with a further question like \emph{so that used up my credit, right?}. If the topic of refund credits has been popped from the stack, this can no longer help clarify what the user wants to know.
Since there is in principle no restriction to how humans revisit and interleave topics in a conversation, we are interested in a more flexible structure than a stack.

\paragraph{Recurrent Neural Networks}
A common choice in recent years has been to use a recurrent neural network (RNN) to process the sequence of previous dialogue turns, both for open domain~\citep{sordoni2015hierarchical, serban2016building} and task-oriented systems~\citep{williams2017hybrid}.
Given enough training data, an RNN should be able to learn any desired behaviour. 
However, in a typical low-resource setting where no large corpus for training a particular task is available, an RNN is not guaranteed to learn to generalize these behaviours. 
Previous work on modifying the basic RNN structure to include inductive biases for this behaviour into a dialogue policy was conducted by Vlasov et al.~\citep{vlasov2018few} and Sahay et al.~\citep{sahay-etal-2019-modeling}. 
These works aim to overcome a feature of RNNs that is undesirable for dialogue modeling. 
RNNs by default consume the \emph{entire} sequence of input elements to produce an encoding, unless a more complex structure like a Long Short-Term Memory (LSTM) cell is trained on sufficient data to explicitly learn that it should `forget' parts of a sequence. 

\paragraph{Transformers}  

The transformer architecture has in recent years replaced recurrent neural networks as the standard for training language models, with models such as Transformer-XL~\citep{dai2019transformer} and GPT-2~\citep{radford2019language} achieving much lower perplexities across a variety of corpora and producing representations that are useful for a variety of downstream tasks~\citep{wang2018glue, devlin2018bert}.
In addition, transformers have recently shown to be more robust to unexpected inputs (such as adversarial examples)~\citep{hsieh2019robustness}.
Intuitively, because the self-attention mechanism preselects which tokens will contribute to the current state of the encoder, a transformer can ignore uninformative (or adversarial) tokens in a sequence. 
To make a prediction at each time step, an LSTM needs to update its internal memory cell, propagating this update to further time steps. 
If an input at the current time step is unexpected, the internal state gets perturbed and at the next time step the neural network encounters a memory state unlike anything encountered during training. 
A transformer accounts for time history via a self-attention mechanism, making the predictions at each time step independent of each other. 
If a transformer receives an irrelevant input, it can ignore it and use only the relevant previous inputs to make a prediction.

Since a transformer chooses which elements in a sequence to use to produce an encoder state at every step, we hypothesise that it could be a useful architecture for processing dialogue histories. 
The sequence of utterances in a conversation may represent multiple interleaved topics, and the transformer's self-attention mechanism can simultaneously learn to disentangle these discourse segments and also to respond appropriately. 

\section{Related Work}

\paragraph{Transformers for open-domain dialogue}

Multiple authors have recently used transformer architectures in dialogue modeling.
Henderson et al.~\citep{henderson2019training} train response selection models on a large dataset from Reddit where both
the dialogue context and responses are encoded with a transformer. 
They show that these architectures can be pre-trained on a large, diverse dataset and later fine-tuned for task-oriented dialogue in specific domains.
Dinan et al.~\citep{dinan2018wizard} used a similar approach, using tranformers to encode the dialogue context as well as
background knowledge for studying grounded open-domain conversations. 
Their proposed architecture comes in two forms: a retrieval model where another transformer is used to
encode candidate responses which are selected by ranking, and a generative model where a transformer is
used as a decoder to produce responses token-by-token.
The key difference with these approaches is that we apply self-attention at the discourse level,
attending over the sequence of dialogue turns rather than the sequence of tokens in a single turn.

\paragraph{Topic disentanglement in task-oriented dialogue}

Recent work has attempted to produce neural architectures for dialogue policies which can handle interleaved discourse segments in a single conversation. 
Vlasov et al.~\citep{vlasov2018few} introduced the Recurrent Embedding Dialogue Policy (REDP) architecture.
The ablation study in this work highlighted that the improved performance of REDP is due to an attention
mechanism over the dialogue history and a copy mechanism to recover from unexpected user input.
This modification to the standard RNN structure enables the dialogue policy to `skip' specific turns in the dialogue history and produce an encoder state which is identical before and after the unexpected input. 
Sahay et al.~\citep{sahay-etal-2019-modeling} develop this line of investigation further by studying the effectiveness of different attention mechanisms for learning this masking behaviour. 

In this work we do not augment the basic RNN architecture but rather replace it with a transformer.
By default, an RNN processes every item in a sequence to calculate an encoding.
REDP's modifications were motivated by the fact that not all dialogue history is relevant.  
Taking this line of reasoning further, we can use self-attention in place of an RNN, so there is no
\emph{a priori} assumption that the whole sequence is relevant, but rather that the dialogue policy should select which historical turns are relevant for choosing a response.

\section{Transformer as a dialogue policy}
\label{sec:tedp}

We propose the Transformer Embedding Dialogue (TED) policy, which greatly simplifies the architecture of the REDP.
Similar to the REDP, we do not use a classifier to select a system action.
Instead, we jointly train embeddings for the dialogue state and each of the system actions by maximizing a similarity function between them. 
At inference time, the current state of the dialogue is compared to all possible system actions, and the one with the highest similarity is selected.
A similar approach is taken by~\citep{bordes2016learning, mehri2019pretraining, henderson2019training} in training retrieval models for task-oriented dialogue.
    
\begin{figure}
\centering
    \includegraphics[width=\linewidth]{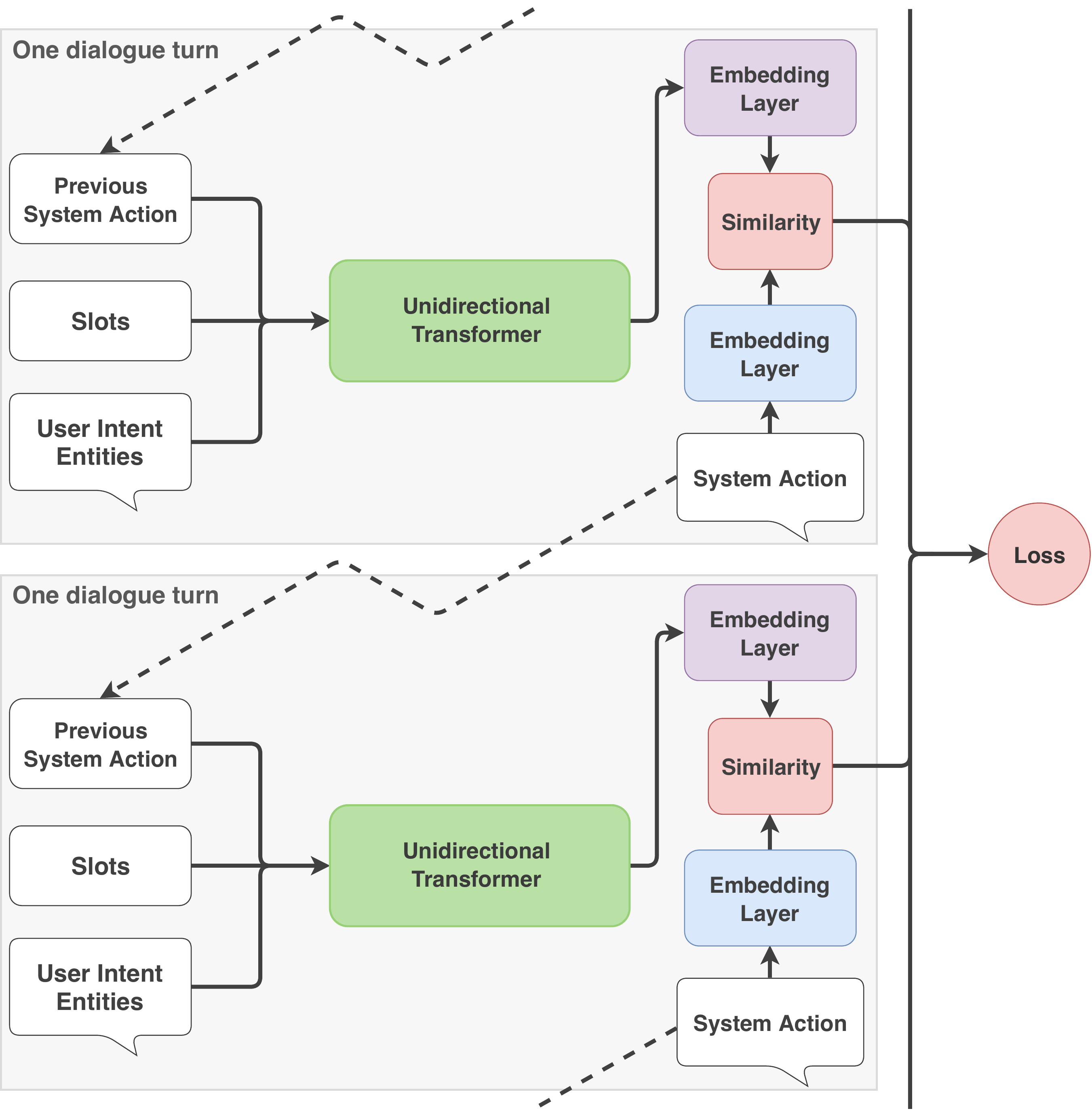}
    \caption{A schematic representation of two time steps of the transformer embedding dialogue policy.}
    \label{fig:tedp}
\end{figure}
Two time steps (i.e. dialogue turns) of the TED policy are illustrated in Figure~\ref{fig:tedp}. A step consists of several key parts.

\paragraph{Featurization}
Firstly, the policy featurizes the user input, system actions and slots.

The TED policy can be used in an end-to-end or in a modular fashion. 
The modular approach is similar to that taken in POMDP-based dialogue policies \citep{williams2007partially} or Hybrid Code Networks~\citep{williams2017hybrid, bocklisch2017rasa}.
An external natural language understanding system is used and the user input 
is featurized as a binary vector indicating the recognized intent and the detected entities.
The dialogue policy predicts an action from a fixed list of system actions. System actions are featurized as binary vectors representing the action name, following the REDP approach
explained in detail in ~\citep{vlasov2018few}.

By end-to-end we mean that there is no supervision beyond the sequence of utterances.
That is, there are no gold labels for the NLU output or the system action names. 
The end-to-end TED policy is still a retrieval model and does not generate new responses.
In the end-to-end setup, user and system utterances are encoded as bag-of-words vectors.

Slots are always featurized as binary vectors, indicating their presence, absence, or that the value is not important to the user, at each step of the dialogue. 
We use a simple slot tracking method, overwriting each slot with the most recently specified value. 

\paragraph{Transformer}
The input to the transformer is the sequence of user inputs and system actions.
Therefore, we leverage the self-attention mechanism present in the transformer to access different parts of dialogue history dynamically at each dialogue turn.
The relevance of previous dialogue turns is learned from data and calculated anew at each
turn in the dialogue.
Crucially, this allows the dialogue policy to take a user utterance into account at one turn
but ignore it completely at another.  

\paragraph{Similarity}
The transformer output $a_{\text{dialogue}}$ and system actions $y_{\text{action}}$ are embedded into a single semantic vector space $h_{\text{dialogue}} = E(a_{\text{dialogue}})$,  $h_{\text{action}} = E(y_{\text{action}})$, where $h\in{\rm I\!R}^{20}$. We use the dot-product loss~\citep{wu2017starspace, henderson2019training} to maximize the similarity $S^+ = h_{\text{dialogue}}^T h_{\text{action}}^+$ with the target label $y_{\text{action}}^+$ and minimize similarities $S^- = h_{\text{dialogue}}^T h_{\text{action}}^-$ with negative samples $y_{\text{action}}^-$. Thus, the loss function for one dialogue reads
\begin{equation}
    L_{\text{dialogue}} = - \biggl\langle S^+ - \log\biggl(e^{S^+} + \sum_{\Omega^-}e^{S^-}\biggr) \biggr\rangle,
    \label{eq:intent}
\end{equation}
where the sum is taken over the set of negative samples $\Omega^-$ and the average $\langle . \rangle$ is taken over time steps inside one dialogue.

The global loss is an average of all loss functions from all dialogues.

At inference time, the dot-product similarity serves as a ranker for the next utterance retrieval problem.

During modular training, we use a balanced batching strategy %\footnote{Balanced batching was introduced in Rasa 1.3.}
to mitigate class imbalance, as some system actions are far more frequent than others.

\section{Experiments}

The aim of our experiments is to compare the performance of the transformer against that of an LSTM on multi-turn conversations. 
Specifically, we want to test the TED policy on the task of picking out relevant turns in the dialogue history for next action prediction.
Therefore, we need a conversational dataset for which system actions depend on the dialogue history across several turns.
This requirement precludes question-answering datasets such as WikiQA~\cite{yangWikiQAChallengeDataset2015} as candidates for evaluation.

In addition, system actions need to be labeled to evaluate next action retrieval accuracy.
Note, that metrics such as Recall@k \cite{loweUbuntuDialogueCorpus2016} could be used on unlabeled data, but since typical dialogues contain many generic responses, such as ``yes'', that are correct in a large number of situations, the meaningfulness of Recall@k is questionable.
We therefore exclude unlabeled dialogue corpora such as the Ubuntu Dialogue Corpus~\cite{loweUbuntuDialogueCorpus2016} or MetalWOZ~\cite{MetaLWOz} from our experiments.

% Wikipedia Question-Answer Dataset
% Yahoo Language Data
% TREC QA Collection
% Relational Strategies in Customer Service Dataset
% Customer Support on Twitter
% Semantic Web Interest Group IRC Chat Logs
% Cornell Movie-Dialogs Corpus
% ConvAI2 Dataset
% Santa Barbara Corpus of Spoken American English
% The NPS Chat Corpus
% Maluuba Goal-Oriented Dialogue
% DSTC
To our knowledge, the only publicly available dialogue datasets that might satisfy both our criteria are the REDP dataset \cite{vlasov2018few}, MultiWOZ~\cite{budzianowski2018multiwoz, eric2019multiwoz} and Google Taskmaster-1~\cite{byrneTaskmaster1RealisticDiverse2019}.
For the latter, we would have to extract action labels from the entity annotations, which is not always possible.
%Unfortunately, the latter two datasets are actually nearly history independent, which we demonstrate in \S\ref{sec:multiwoz}, before we evaluate the TED policy on the remaining dataset~\cite{vlasov2018few} in \S\ref{sec:subdial}.

Two different models serve as baseline in our experiments. 
First, the REDP model by~\citet{vlasov2018few}, which was specifically designed to handle long-range history dependencies, but is LSTM-based. 
Second, another LSTM-based policy that is identical to TED, except that the transformer was replaced by an LSTM.

We use the first (REDP) baseline for the experiments on the \citep{vlasov2018few} dataset, as this baseline is stronger when long-range dependencies are in play.
For the MultiWOZ experiments, we only compare to the simple LSTM policy, since the MultiWOZ dataset is nearly history independent as we demonstrate here.
 
%All experiments are performed using the Rasa framework~\citep{bocklisch2017rasa} and all code, hyperparameters, and data required to reproduce
All experiments are available online at \url{https://github.com/RasaHQ/TED-paper}.

\subsection{Conversations containing sub-dialogues}
\label{sec:subdial}

    We first evaluate experiments on the dataset of~\citet{vlasov2018few}.
    This dataset was specifically designed to test the ability of a dialogue policy to handle non-cooperative or unexpected user input. It consists of task-oriented dialogues in hotel and restaurant reservation domains containing cooperative (user provides necessary information related to the task) and non-cooperative (user asks a question unrelated to the task or makes chit-chat) dialogue turns.
    One of the properties of this dataset is that the system repeats the previously asked question after any non-cooperative user behavior.
    This dataset is also used in~\citep{sahay-etal-2019-modeling} to compare the performance of different attention mechanisms.

    \begin{figure}
        \centering
        \includegraphics[width=\linewidth]{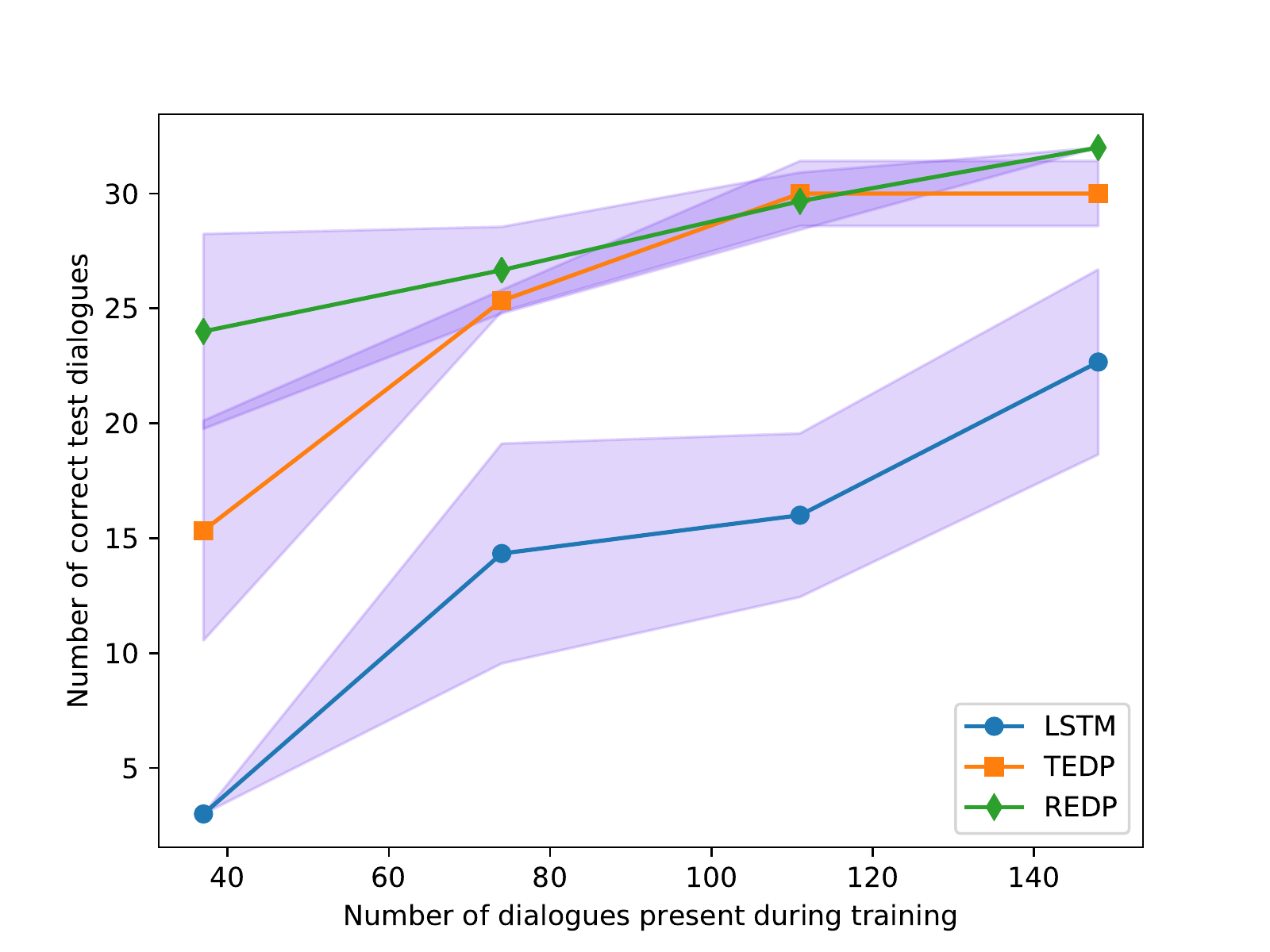}
        \caption{Performance of REDP, TED policy, and an LSTM baseline policy on the dataset from~\citep{vlasov2018few}. Lines indicate mean and shaded area the standard deviation of 3 runs. The horizontal axis indicates the amount of training conversations used to train the model and the vertical axis is the number of conversations in the test set in which every action is predicted correctly.}
        \label{fig:compare_models}
    \end{figure}

    Figure~\ref{fig:compare_models} shows the performance of different dialogue policies on the held-out test dialogues as a function of the amount of conversations used to train the model. The TED policy performs on par with REDP without any specifically designed architecture to solve the task and significantly outperforms a simple LSTM-based policy. 
    In the extreme low-data regime, the TED policy is outperformed by REDP. It should be noted that REDP relies heavily on its copy mechanism to predict the previously asked question after a non-cooperative digression.
    However, the TED policy, being both simpler and more general, achieves similar performance without relying on dialogue properties like repeating a question. 
    Moreover, due to the transformer architecture, the TED policy trains faster than REDP and requires fewer training epochs to achieve the same accuracy.
    
    \begin{figure}
        \includegraphics[width=\linewidth]{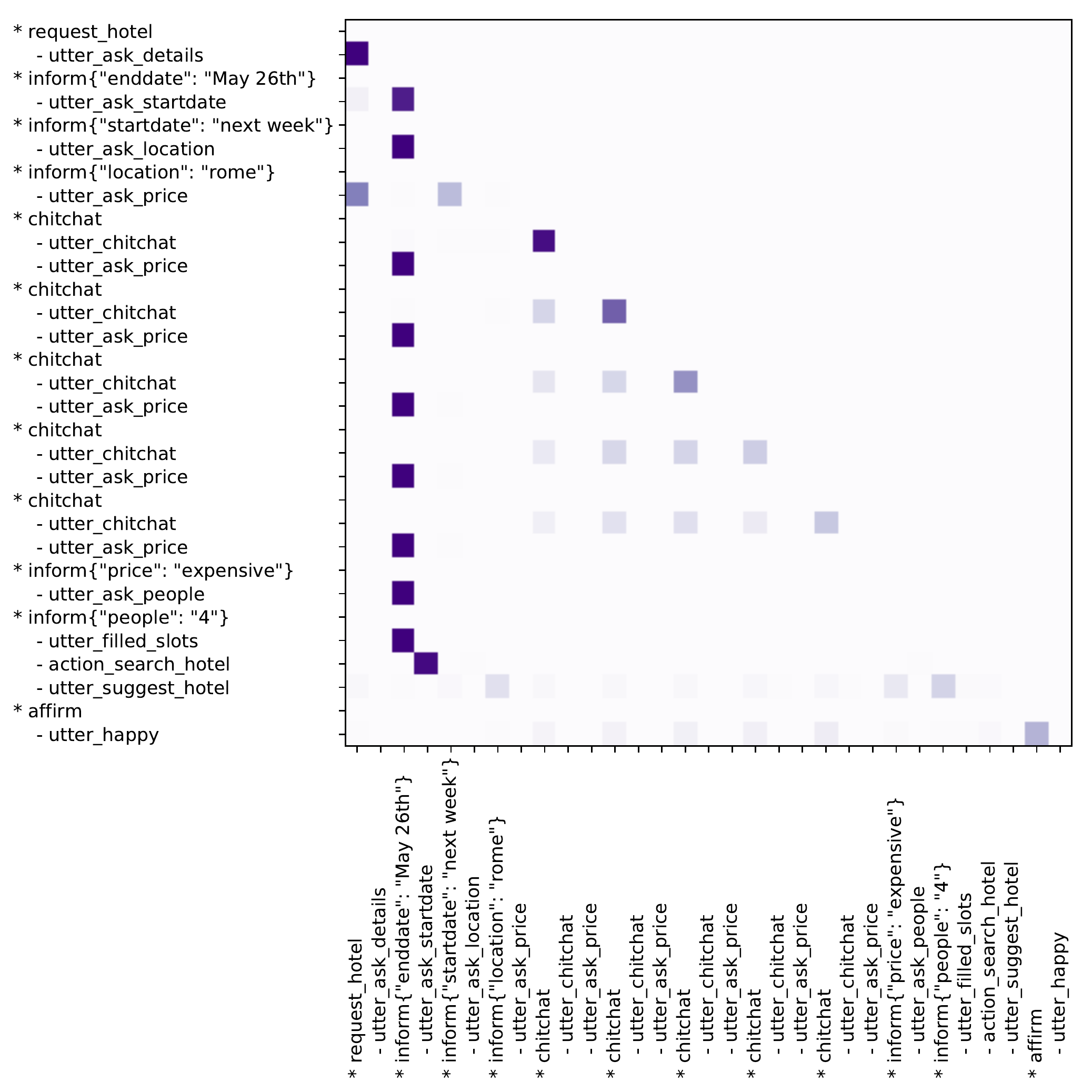}
        \caption{Attention weights of the TED policy on an example dialogue. On the vertical axis are predicted dialogue turns, and on the horizontal axis is the dialogue history over which the TED policy attends. We use a unidirectional transformer, so the upper triangle is masked with zeros to avoid attending to future dialogue turns.}
        \label{fig:attn}
    \end{figure}
    
    Figure~\ref{fig:attn} visualizes the attention weights of the TED policy on an example dialogue.
    This example dialogue contains several chit-chat utterances in a row in the middle of the conversation. 
    The Figure shows that the series of chit-chat interactions is completely ignored by the self-attention mechanism when task completion is attempted (i.e.\ further required questions are asked).
    Note, that the learned weights are sparse, even though the TED policy does not use a sparse attention architecture. 
    Importantly, the TED policy chooses key dialogue steps from the history that are relevant for the current prediction and ignores uninformative history.
    Here, we visualize only one conversation, but the result is the same for an arbitrary number of chit-chat dialogue turns.

\subsection{Comparing the end-to-end and modular approaches on MultiWOZ}
\label{sec:multiwoz}

    Having demonstrated that the light-weight TED policy performs at least on par with the specialized REDP and significantly outperforms a basic LSTM policy when evaluated on conversations that contain long-range history dependencies, we now compare TED to an LSTM policy on the MultiWOZ 2.1 dataset.
    In contrast to the previous Section, the LSTM policy of the present Section is an architecture identical to TED, but with the transformer replaced by an LSTM cell.
    
    We chose MultiWOZ for this experiment because it concerns multi-turn conversations and provides system action labels.
    %To the best of our knowledge, the only other large, publicly available dialogue dataset besides MultiWOZ and the REDP dataset we use in \S\ref{sec:subdial} that satisfies both of these criteria is Google Taskmaster-1~\cite{byrneTaskmaster1RealisticDiverse2019}.
    % For the latter, we would have to extract action labels from their the annotations, which is not always possible.
    % In addition, system actions need to be labeled to evaluate next action retrieval accuracy.
    Unfortunately, we discovered that it does not contain many long-range dependencies, as we shall demonstrate later in this Section.
    Therefore, neither TED nor the REDP have any conceptual advantages over an LSTM.
    Subsequently we show that the TED policy performs on par with an LSTM on this commonly used benchmark dataset.
    
    MultiWOZ 2.1 is a dataset of 10438 human-human dialogues for a Wizard-of-Oz task in seven different domains: hotel, restaurant, train, taxi, attraction, hospital, and police.
    In particular, the dialogues are between a user and a clerk (wizard).
    The user asks for information and the wizard, who has access to a knowledge base about all the possible things that the user may ask for, provides that information or executes a booking.
    The dialogues are annotated with labels for the wizard's actions, as well as the wizard's knowledge about the user's goal after each user turn.
    %See Figure~\ref{fig:compare_SNG0253} for an example.
    
    For our experiments, we split the MultiWOZ 2.1 dataset into a training and a test set of $7249$ and $1812$ dialogues, respectively.
    Unfortunately, we had to neglect $1377$ dialogues altogether, since their annotations are incomplete.

    \paragraph{End-to-end training.}

    As a first experiment on MultiWOZ 2.1 we study an end-to-end retrieval setup, where the user utterance is used directly as input to the TED policy, which then has to retrieve the correct response from a predefined list (extracted from MultiWOZ).
    
    The wizard's behaviour depends on the result of queries to the knowledge base. For example, if only a single venue is returned, the wizard will probably refer to it.
    We marginalize this knowledge base dependence by (i) delexicalizing all user and wizard utterances \citep{mrkvsic2016neural}, and (ii) introducing status slots that indicate whether a venue is available, not available, already booked, or unique (i.e.\ the wizard is going to recommend or book a particular venue in the next turn). 
    These slots are featurized as a 1-of-K binary vector.

    To compute the accuracy and F1 scores of the TED policy's predictions, we assign the action labels (e.g. \verb+request_restaurant+) that are provided by the MultiWOZ dataset to the output utterances, and compare them to the correct labels.
    If multiple labels are present, we concatenate them in alphabetic order to a single label. 

    Table~\ref{tab:multiwoz} shows the resulting F1 scores and accuracies on the hold-out test set.
    The discrepancy between the F1 score and the accuracy stems from the fact that some labels, s.a.\ \verb+bye_general+, occur very frequently (4759 times) compared to most other labels, s.a.\ \verb+recommend_restaurant_select_restau+ \verb+rant+, which only occurs 11 times.

    The fact that accuracy and F1 scores are generally low compared to 1.0 stems from a deeper issue with the MultiWOZ dialog dataset.
    Specifically, because more than one particular behaviour of the wizard would be considered 'correct' in most situations, the MultiWOZ dataset is unsuitable for supervised learning of dialogue policies.
    Put differently, some of the wizards' actions in MultiWOZ are not deterministic, but probabilistic.
    For example, it cannot be learned when the wizard should ask if the user needs anything else, since this is the personal preference of the people who take the wizard's role.
    We elaborate on this and several other issues of the MultiWOZ dataset in~\citep{mosig2020context}.
    
    \begin{table}
        \centering
        \begin{tabular}{|llll|}
        \hline
        \textbf{model} & \textbf{$N$} & \textbf{accuracy} & \textbf{F1 score} \\
        \hline\hline
        TED end-to-end & 10 & 0.64 & 0.28 \\  % ok . % updated to use all stories
         & 2 & 0.62 & 0.24 \\  % ok % updated to use all stories
        \hline
        TED modular & 10 & 0.73 & 0.63 \\ % ok % updated to use all stories
         & 2 & 0.69 & 0.55 \\ % ok % updated to use all stories
        \hline
        LSTM end-to-end & 10 & 0.51 & 0.23 \\ 
        %LSTM end-to-end & 10 & 0.506 & 0.234 \\ 
         & 2 & 0.57 & 0.24 \\   % 17h27m
        %  & 2 & 0.568 & 0.240 \\ 
        \hline
        LSTM modular & 10 & 0.68 & 0.60 \\ 
        %LSTM modular & 10 & 0.681 & 0.595 \\ 
         & 2 & 0.61 & 0.54 \\ 
        % & 2 & 0.614 & 0.538 \\
        \hline
        % end-to-end & 10 & 0.641 & 0.276 \\  % ok . % updated to use all stories
        % (with status slots) & 2 & 0.622 & 0.238 \\  % ok % updated to use all stories
        % \hline
        % modular & 10 & 0.726 & 0.634 \\ % ok % updated to use all stories
        % (with status slots) & 2 & 0.689 & 0.553 \\ % ok % updated to use all stories
        % \hline
        \end{tabular}
        \caption{%
        Accuracy and F1 scores of the TED policy in end-to-end and modular mode,
        as well as the TED policy with the transformer replaced by an LSTM. 
        Models are evaluated on the MultiWOZ 2.1 dataset using max\_history $N$.
        All scores concern prediction at the action level on the test set.
        %While the modular architecture generally achieves higher scores, taking into account more than 2 steps in the dialogue history barely increases the performance in either case.% updated to use all stories
        }
        \label{tab:multiwoz}
    \end{table}

    % Comparing row 'end-to-end' to row 'end-to-end (with status slots)' with \verb+max_history = 10+ in Table~\ref{tab:multiwoz}, we see that introducing status slots raises the accuracy by $0.035$, and the F1 score by $0.068$, but scores remain low compared to 1.0. % updated to use all stories
    % We shall explore the reason for these low scores in conjunction with analysing the results of the modular architecture predictions in the next section.
    
    \paragraph{Modular training.}

    We now repeat the above experiment, using the same subset of MultiWOZ dialogues, but now adopting the modular approach.
    We simulate an external natural language understanding pipeline and provide gold user intents and entities to the TED policy instead of the original user utterances.  
    We extract the intents from the changes in the Wizard's belief state.
    This belief state is provided by the MultiWOZ dataset in the form of a set of slots (e.g. \verb+restaurant_area+, \verb+hotel_name+, etc.) that get updated after each user turn.
    A typical user intent is thus \verb+inform{"restaurant_area": "south"}+.
    The user does not always provide new information, however, so the intent might be simply \verb+inform+ (without any entities).
    If the last user intent of the dialogue was uninformative in this way, we assume it is a farewell and thus annotate it as \verb+bye+.
    
    Using the modular approach instead of end-to-end learning roughly doubles the F1 score and also increases accuracy slightly, as can be seen in Table~\ref{tab:multiwoz}. 
    This is not surprising since the modular approach receives additional supervision.
    
\begin{figure*}
        \begin{center}
        \includegraphics[width=0.8\linewidth]{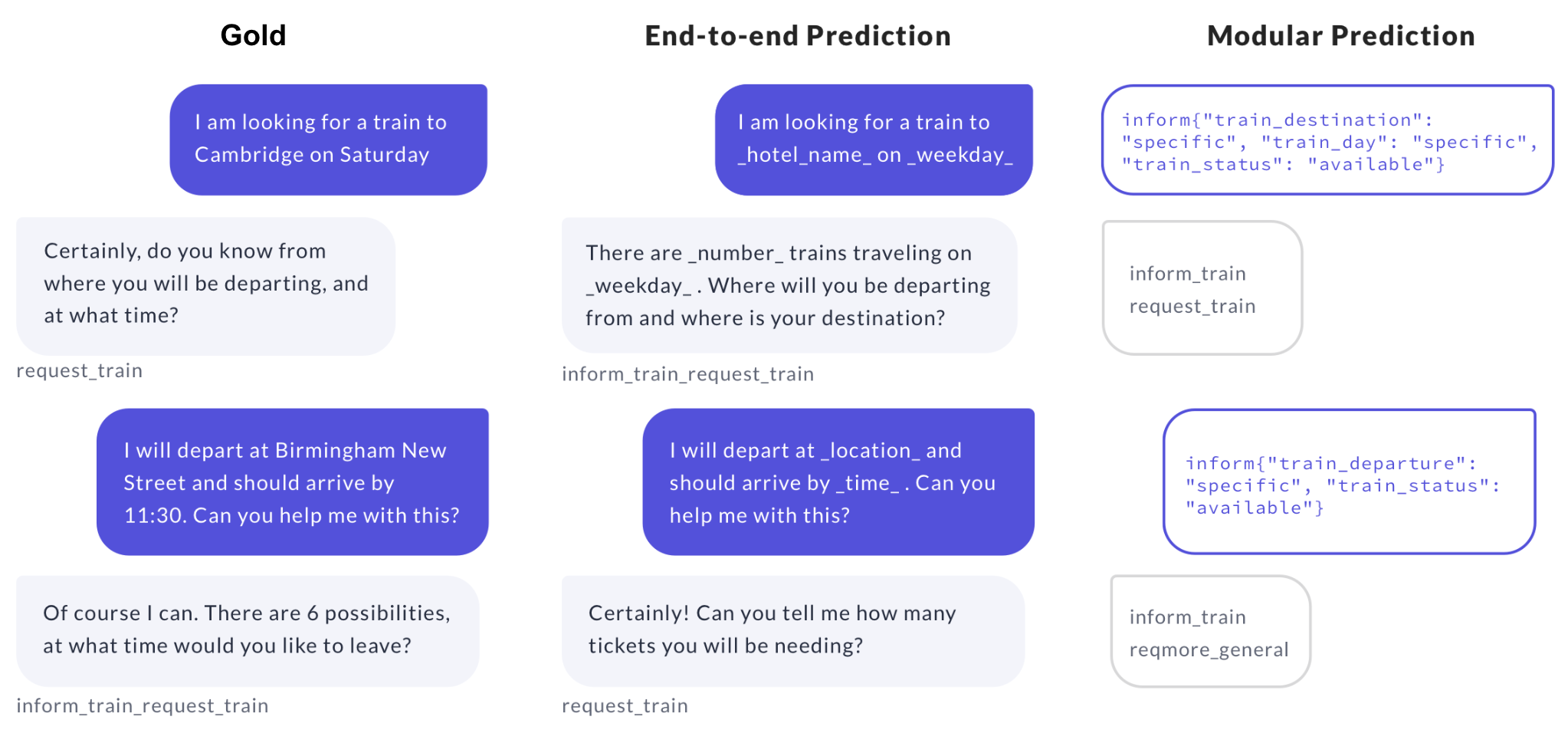}
        \caption{MultiWOZ 2.1 dialogue SNG0253, as-is (first column), as predicted by the end-to-end TED policy (second column), and as sequence of user intents and system actions predicted by modular TED policy (third column). The two predictions are sensible, yet incorrect according to the labels. Furthermore, the end-to-end and modular TED policies make similar kinds of mistakes.}
        \label{fig:compare_SNG0253}
        \end{center}
    \end{figure*}

    Although the scores suggest that the modular TED policy performs better than the end-to-end TED policy, the \textit{kinds} of mistakes made are similar.
    We demonstrate this with one example dialog from our test set, named \verb+SNG0253+, that is displayed in Figure~\ref{fig:compare_SNG0253}.
    
    The second column of Figure~\ref{fig:compare_SNG0253} shows the end-to-end predictions.
    The two predicted responses are both sensible, i.e.\ the replies could have come from a human.
    Nevertheless, both results are marked as wrong, because according to the gold dialogue (first column), the first response should only have included its second sentence (\verb+request_train+, but not \verb+inform_train+).
    For the fourth turn, however, it is the other way around: According to the target dialogue the response should have included additional information about the train (\verb+inform_train_request_train+), whereas the predicted dialogue only asked for more information (\verb+request_train+).
   
    The third column shows that the modular TED policy makes the same kinds of mistakes: instead of predicting only \verb+request_train+, it predicts to take both actions, \verb+inform_train+ and \verb+request_train+ in the second turn.
    In the final turn, instead of \verb+request_train+, the modular TED policy predicts \verb+reqmore_general+, which means that the wizard asks if the user requires anything else.
    This reply is perfectly sensible and does in fact occur in similar dialogues of the training set (see, e.g., Dialogue \verb+PMUL1883+).  
    Thus, \textit{the} correct behaviour doesn't exist and it is impossible to achieve high scores, as reflected by the test scores of Table~\ref{tab:multiwoz}.
    
    To the best of our knowledge, the state of the art F1 scores on next action retrieval with MultiWOZ are given by~\citep{mehri2019pretraining} and~\citep{mehri2019multi} with $0.64$ and $0.72$, respectively. However, these numbers are not directly comparable to ours: we retrieve actions out of all $56128$ possible responses and compare the label of the retrieved response with the label of correct response, while they retrieve out of $20$ negative samples and compare text responses directly.
    
    \paragraph{History independence.}
    
    As Table~\ref{tab:multiwoz} shows, taking into account only the last two turns (i.e.\ the current user utterance or intent, and one system action before that), instead of the last 10 turns, the accuracy and F1 scores decrease by no more than $0.04$ for end-to-end and no more than $0.08$ for the modular architecture.
    For the end-to-end LSTM architecture that we discuss in the next paragraph, the performance even improves when less history is taken into account.
    Thus, MultiWOZ appears to depend only weakly on the dialogue history, and therefore we cannot evaluate how well the TED policy handles dialogue complexity.
    
    \paragraph{Transformer vs LSTM.}
    
    As a final experiment, we replace the transformer in the TED architecture by an LSTM and run the same experiments as before.
    The results are shown in Table~\ref{tab:multiwoz}.
    
    The F1 scores of the LSTM and transformer versions differ by no more than $0.05$, which is to be expected, since in MultiWOZ the vast majority of information is carried by the most recent turn.
    
    The LSTM version lacks the accuracy of the transformer version, however.
    Specifically, the accuracy scores for end-to-end training are up to $0.13$ points lower for the LSTM.
    It is difficult to assert the reason for this discrepancy due to the problems of ambiguity that we have identified earlier in this Section.

\section{Conclusions}

We introduce the transformer embedding dialogue (TED) policy in which a transformer's self-attention mechanism operates over the sequence of dialogue turns.
We argue that this is a more appropriate architecture than an RNN due to the presence of interleaved topics in real-life conversations.
We show that the TED policy can be applied to the MultiWOZ dataset in both a modular and end-to-end fashion, although we also find that this dataset is not ideal for supervised learning of dialogue policies, due to a lack of history dependence and a dependence on individual crowd-worker preferences.
We also perform experiments on a task-oriented dataset specifically created to test the ability to recover from non-cooperative user behaviour. 
The TED policy outperforms the baseline LSTM approach and performs on par with REDP, despite TED being faster, simpler, and more general. We demonstrate that learned attention weights are easily interpretable and reflect dialogue logic. At every dialogue turn, a transformer picks which previous turns to take into account for current prediction, selectively ignoring or attending to different turns of the dialogue history.

\subsection*{Acknowledgments}

We would like to thank the Rasa team and Rasa community for feedback and support. Special thanks to Elise Boyd for supporting us with the illustrations.

\bibliography{core_transformer}

\end{document}